\documentclass[10pt,twocolumn]{article}

\usepackage[margin=1in]{geometry}
\usepackage{graphicx}
\usepackage{amsmath}
\usepackage{amssymb}
\usepackage{booktabs}
\usepackage{hyperref}
\usepackage{authblk}
\usepackage{times}

\title{\textbf{Evaluating Model-Agnostic Meta-Learning on MetaWorld ML10 Benchmark: Fast Adaptation in Robotic Manipulation Tasks}}

\author{Sanjar Atamuradov}
\affil{Georgia Institute of Technology, Atlanta, GA\\
\texttt{satamuradov3@gatech.edu}}

\date{}

\begin{document}

\maketitle

\begin{abstract}
Meta-learning algorithms enable rapid adaptation to new tasks with minimal data, a critical capability for real-world robotic systems. This paper evaluates Model-Agnostic Meta-Learning (MAML) combined with Trust Region Policy Optimization (TRPO) on the MetaWorld ML10 benchmark, a challenging suite of ten diverse robotic manipulation tasks. We implement and analyze MAML-TRPO's ability to learn a universal initialization that facilitates few-shot adaptation across semantically different manipulation behaviors including pushing, picking, and drawer manipulation. Our experiments demonstrate that MAML achieves effective one-shot adaptation with clear performance improvements after a single gradient update, reaching final success rates of 21.0\% on training tasks and 13.2\% on held-out test tasks. However, we observe a generalization gap that emerges during meta-training, where performance on test tasks plateaus while training task performance continues to improve. Task-level analysis reveals high variance in adaptation effectiveness, with success rates ranging from 0\% to 80\% across different manipulation skills. These findings highlight both the promise and current limitations of gradient-based meta-learning for diverse robotic manipulation, and suggest directions for future work in task-aware adaptation and structured policy architectures.
\end{abstract}


\section{Introduction}

Meta-learning, or ``learning to learn,'' represents a paradigm shift in machine learning from task-specific optimization to learning algorithms that can rapidly adapt across task distributions \cite{finn2017model}. Rather than training models from scratch for each new task, meta-learning algorithms optimize for transferable knowledge that enables quick adaptation with minimal data. This capability is particularly valuable in robotics, where collecting training data is expensive and robots must generalize across diverse scenarios.

Model-Agnostic Meta-Learning (MAML) \cite{finn2017model} introduced a simple yet powerful approach: optimize a model's initial parameters such that a few gradient steps on a new task lead to effective performance. MAML employs a bi-level optimization strategy with an inner loop that adapts to individual tasks and an outer loop that updates the initialization to improve adaptation across the task distribution. Critically, MAML is model-agnostic—it can be applied to any gradient-based learning algorithm, spanning supervised learning, reinforcement learning, and beyond.

In the reinforcement learning (RL) domain, MAML has demonstrated promise on locomotion tasks in simulated environments such as MuJoCo \cite{finn2017model}. However, these early benchmarks primarily involved variations in dynamics or goal positions within similar task families (e.g., different running directions for a simulated robot). Real-world robotic applications demand adaptation across more diverse and complex behaviors.

This work evaluates MAML-TRPO on the MetaWorld ML10 benchmark \cite{yu2019meta}, which consists of ten distinct robotic manipulation tasks with different action requirements, reward structures, and semantic meanings. Tasks include button pressing, drawer opening, door manipulation, and object placement—each requiring different control strategies and mechanical understanding. This benchmark provides a more rigorous test of meta-learning's ability to discover initializations that generalize across meaningfully different skills rather than minor task variations.

Our contributions are:
\begin{itemize}
    \item Implementation and evaluation of MAML-TRPO on the MetaWorld ML10 benchmark for multi-task robotic manipulation
    \item Analysis of adaptation dynamics showing effective one-shot learning with clear pre- and post-adaptation performance gaps
    \item Identification of a generalization gap where test task performance plateaus while training performance improves
    \item Task-level breakdown revealing high variance in adaptation success across manipulation types
    \item Discussion of limitations and future directions for gradient-based meta-RL in robotics
\end{itemize}

\section{Related Work}

\subsection{Meta-Learning}

Meta-learning has emerged as a unifying framework for learning algorithms that improve with experience across multiple tasks. Early approaches included learning-to-learn via gradient descent \cite{ravi2017optimization} and memory-augmented neural networks. MAML \cite{finn2017model} distinguished itself through its model-agnostic formulation and applicability to both supervised and reinforcement learning settings.

Subsequent work has extended MAML in various directions. \textit{How to Train Your MAML} \cite{antoniou2019train} introduced techniques to stabilize training and improve convergence. Hospedales et al. \cite{hospedales2020meta} provide a comprehensive survey of meta-learning approaches across neural network architectures. In the RL domain, alternative meta-learning strategies have been proposed, including context-based approaches like PEARL \cite{rakelly2019efficient} that use probabilistic inference for task identification, and recurrent methods like RL$^2$ that treat adaptation as a sequential decision problem.

\subsection{Multi-Task Reinforcement Learning}

Multi-task RL aims to train agents that can perform well across multiple tasks, either through shared representations or modular architectures. Approaches include soft modularization \cite{pong2019multi}, which learns task-specific module compositions, and hierarchical RL methods that decompose complex tasks into reusable skills.

The MetaWorld benchmark \cite{yu2019meta} was specifically designed to evaluate meta-learning and multi-task RL algorithms on diverse manipulation tasks. It provides standardized environments with controlled complexity, enabling systematic comparison of different approaches. Our work leverages the ML10 subset, which contains ten distinct manipulation primitives, to assess MAML's cross-task generalization capabilities.

\subsection{Trust Region Policy Optimization}

Trust Region Policy Optimization (TRPO) \cite{schulman2015trust} is a policy gradient method that constrains policy updates to remain within a trust region, ensuring monotonic improvement and training stability. TRPO's robustness makes it well-suited for meta-learning settings where the same algorithm must perform reliably across diverse task distributions. The combination of MAML and TRPO (MAML-TRPO) has shown strong empirical results on continuous control tasks.

\section{Background}

\subsection{Reinforcement Learning Formulation}

We model each task as a Markov Decision Process (MDP) defined by the tuple $(\mathcal{S}, \mathcal{A}, P, r, \rho_0, \gamma)$, where $\mathcal{S}$ is the state space, $\mathcal{A}$ is the action space, $P: \mathcal{S} \times \mathcal{A} \times \mathcal{S} \to \mathbb{R}$ is the transition probability function, $r: \mathcal{S} \times \mathcal{A} \to \mathbb{R}$ is the reward function, $\rho_0$ is the initial state distribution, and $\gamma \in [0,1)$ is the discount factor.

The agent's behavior is governed by a policy $\pi_\theta: \mathcal{S} \to \mathcal{A}$ parameterized by $\theta$. The objective is to maximize the expected return:
\begin{equation}
    J(\theta) = \mathbb{E}_{\tau \sim \pi_\theta} \left[ \sum_{t=0}^{T} \gamma^t r(s_t, a_t) \right]
\end{equation}
where $\tau = (s_0, a_0, r_0, \ldots, s_T)$ denotes a trajectory.

\subsection{Model-Agnostic Meta-Learning}

MAML \cite{finn2017model} addresses the meta-learning problem: given a distribution over tasks $p(\mathcal{T})$, find an initialization $\theta$ such that fine-tuning on a new task $\mathcal{T}_i$ with a few gradient steps yields good performance.

The algorithm operates through a two-level optimization:

\textbf{Inner loop (task-specific adaptation):}
For each task $\mathcal{T}_i$, compute adapted parameters $\theta_i'$ via one or more gradient steps:
\begin{equation}
    \theta_i' = \theta - \alpha \nabla_\theta \mathcal{L}_{\mathcal{T}_i}(\theta)
\end{equation}
where $\alpha$ is the inner learning rate and $\mathcal{L}_{\mathcal{T}_i}$ is the task-specific loss.

\textbf{Outer loop (meta-optimization):}
Update the initialization $\theta$ to minimize the loss after adaptation:
\begin{equation}
    \theta \leftarrow \theta - \beta \nabla_\theta \sum_{\mathcal{T}_i \sim p(\mathcal{T})} \mathcal{L}_{\mathcal{T}_i}(\theta_i')
\end{equation}
where $\beta$ is the meta learning rate.

In the RL setting, the loss $\mathcal{L}_{\mathcal{T}_i}$ is the negative expected return on task $\mathcal{T}_i$. The inner loop collects trajectories and performs policy gradient updates, while the outer loop updates $\theta$ based on post-adaptation performance.

\subsection{MetaWorld ML10 Benchmark}

MetaWorld \cite{yu2019meta} is a benchmark for meta-RL and multi-task learning consisting of 50 robotic manipulation tasks built on the MuJoCo physics simulator. The ML10 subset contains ten diverse tasks:

\begin{itemize}
    \item \textbf{Training tasks:} button-press-topdown, drawer-close, door-open, peg-insert-side, reach, sweep, basketball, window-open
    \item \textbf{Test tasks:} door-close, drawer-open, lever-pull, shelf-place, sweep-into
\end{itemize}

Each task involves a 7-DOF Sawyer robot arm with a parallel gripper. Observations include joint angles, velocities, end-effector position, and task-specific object/goal positions (typically 39-dimensional). Actions are 4-dimensional: 3D end-effector displacement and gripper control.

Rewards are task-specific but generally dense, based on distance to goal positions. Each task includes a binary success condition (e.g., drawer fully opened, object placed on shelf) used for evaluation.

\section{Method}

\subsection{Implementation Details}

We implement MAML-TRPO on MetaWorld ML10 using the following architecture and hyperparameters:

\textbf{Policy Network:} We use a Gaussian Multi-Layer Perceptron (MLP) policy with two hidden layers of 100 units each, separated by $\tanh$ activations. The policy outputs a mean action vector and a learned diagonal covariance matrix for action sampling. This architecture is consistent with the original MAML paper \cite{finn2017model} for continuous control tasks.

\textbf{Value Function:} A separate MLP value function with two hidden layers of 32 units each and $\tanh$ activations is used to estimate the advantage function for policy gradient computation.

\textbf{Meta-Training:} We train for 300 meta-iterations (epochs), where each iteration samples 20 tasks from the ML10 training set. For each task:
\begin{enumerate}
    \item Collect trajectories using the current policy $\pi_\theta$
    \item Compute one inner-loop gradient update with learning rate $\alpha = 0.1$
    \item Collect post-adaptation trajectories using $\pi_{\theta'}$
    \item Compute meta-gradient based on post-adaptation performance
\end{enumerate}

\textbf{Optimization:} We use TRPO for the meta-optimization step, with a KL-divergence constraint of 0.01. The inner loop uses vanilla policy gradient updates.

\textbf{Sampling:} For each task, we collect 10 episodes per iteration, with a maximum episode length of 150 steps. This yields approximately 1,500 timesteps per task per iteration.

\subsection{Training and Evaluation Protocol}

\textbf{Task Sampling:} During meta-training, we randomly sample tasks from the eight training tasks in ML10. Each task instance includes randomized object and goal positions to increase within-task diversity.

\textbf{Evaluation:} We evaluate on five held-out test tasks: door-close, drawer-open, lever-pull, shelf-place, and sweep-into. For each test task, we:
\begin{enumerate}
    \item Collect initial trajectories with the meta-learned initialization
    \item Perform 1-3 gradient updates
    \item Measure success rate and average return after each update
\end{enumerate}

We compare against an untrained baseline (random initialization) to isolate the benefit of meta-learning.

\textbf{Metrics:} We report:
\begin{itemize}
    \item \textbf{Policy Loss:} Pre- and post-adaptation policy loss during training
    \item \textbf{Success Rate:} Percentage of episodes meeting the task-specific success condition
    \item \textbf{Average Return:} Mean cumulative reward over evaluation episodes
\end{itemize}

\section{Results}

\subsection{Adaptation Effect During Training}

\begin{figure}[h]
    \centering
    \includegraphics[width=\columnwidth]{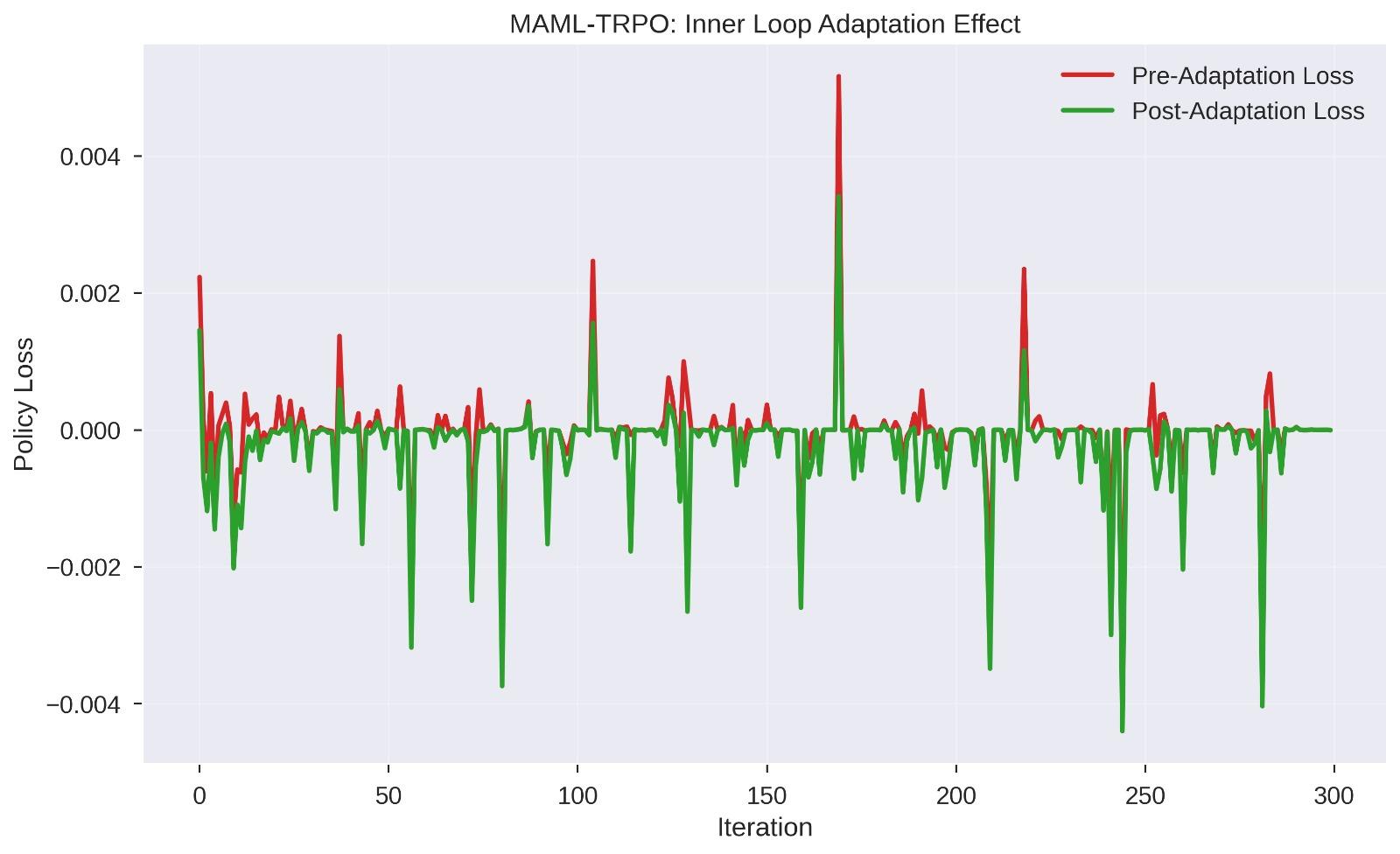}
    \caption{Policy loss before (red) and after (green) adaptation across 300 training iterations on MetaWorld ML10. Post-adaptation consistently improves performance, demonstrating successful few-shot adaptation from the learned initialization.}
    \label{fig:training_loss}
\end{figure}

Figure \ref{fig:training_loss} shows the policy loss before and after one gradient step across 300 meta-training iterations. Each point represents the average loss on a batch of sampled tasks.

Key observations:
\begin{itemize}
    \item \textbf{Pre-adaptation loss} remains above zero throughout training, indicating that the meta-initialized policy does not directly solve new tasks without adaptation
    \item \textbf{Post-adaptation loss} consistently drops below zero after a single gradient step, showing improved expected return
    \item \textbf{Adaptation gap} (difference between pre- and post-adaptation) remains consistent across iterations, confirming that MAML learns an initialization conducive to rapid adaptation
    \item \textbf{Fluctuations} in both curves reflect varying task difficulty within the ML10 benchmark
\end{itemize}

\subsection{Few-Shot Adaptation on Test Tasks}

We evaluate adaptation speed by varying the number of gradient update steps (1-3) on four held-out test tasks, comparing MAML against an untrained baseline.

\begin{figure}[h]
    \centering
    \includegraphics[width=0.5\textwidth]{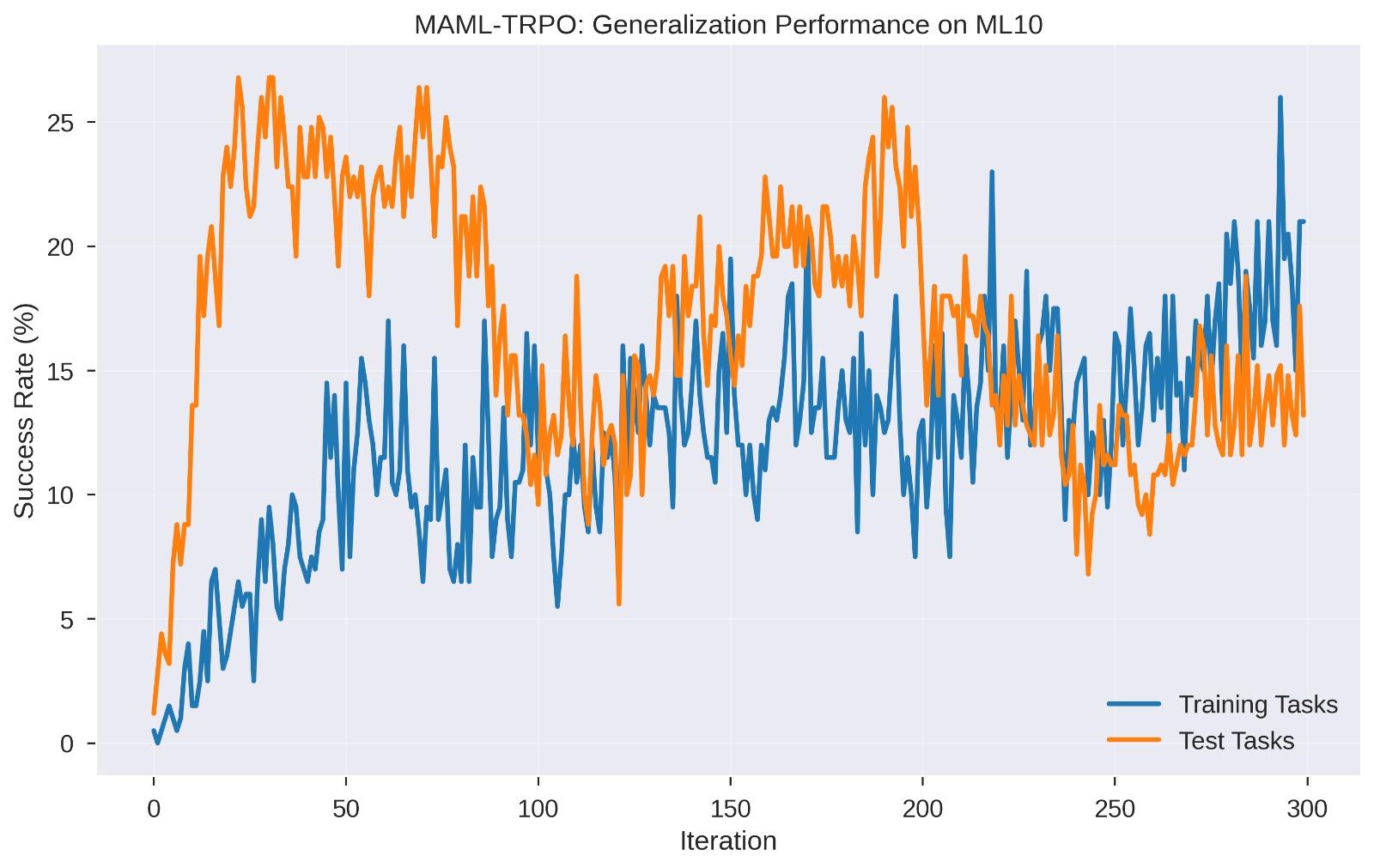}
    \caption{Average return over 1-3 gradient update steps for various test tasks. MAML (blue) compared to untrained baseline (orange). Most tasks show strong one-shot adaptation, but performance degrades with additional steps for some tasks. The figure contains four subplots showing results for: (top left) Lever-pull, (top right) Shelf-place, (bottom left) Door-close, and (bottom right) Drawer-open tasks.}
    \label{fig:gradient_steps}
\end{figure}

Figure \ref{fig:gradient_steps} shows the results:
\begin{itemize}
    \item \textbf{Drawer-open and Shelf-place:} Strong initial adaptation with the first gradient step, but subsequent steps show diminishing or negative returns. This highlights a known MAML limitation: the algorithm optimizes for one-step adaptation, potentially at the expense of multi-step stability.
    \item \textbf{Door-close:} Slower initial adaptation but continued improvement with additional gradient steps, suggesting this task may require more fine-tuning.
    \item \textbf{Lever-pull:} Poor adaptation across all gradient steps, barely outperforming the baseline. This indicates difficulty in transferring manipulation knowledge to tasks requiring sequential or precise control.
\end{itemize}

\subsection{Generalization Dynamics}

Figure \ref{fig:generalization} tracks success rates on training and test tasks throughout meta-training. This reveals an important dynamic:

\begin{figure}[h]
    \centering
    \includegraphics[width=\columnwidth]{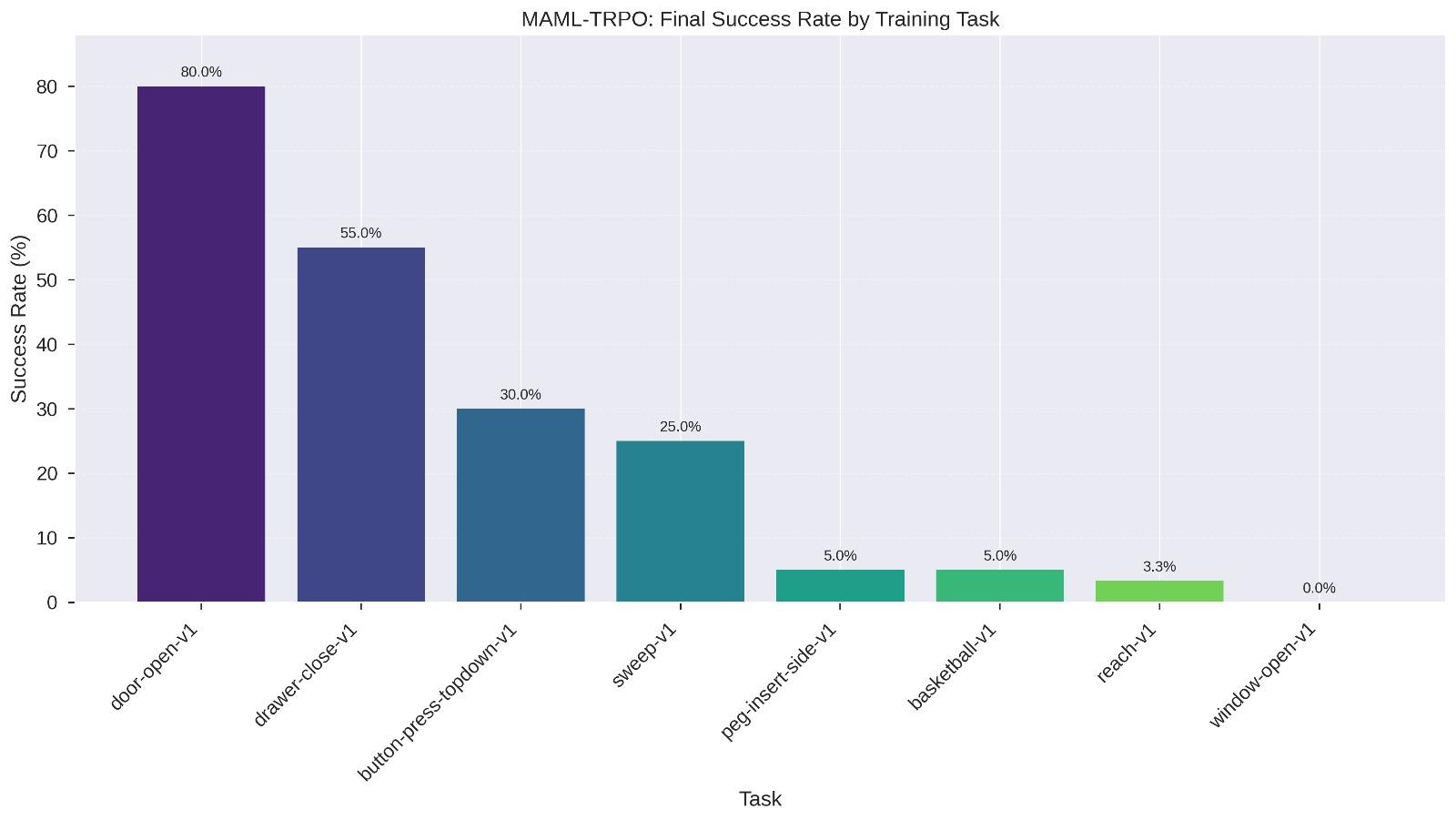}
    \caption{Success rate (\%) on training and test tasks over 300 meta-training iterations. Early generalization to test tasks is overtaken by specialization on training tasks. Final success rates: 21.0\% (training) and 13.2\% (test).}
    \label{fig:generalization}
\end{figure}

\begin{itemize}
    \item \textbf{Early training:} Test task performance initially exceeds training performance, suggesting that early meta-initializations generalize broadly before specializing
    \item \textbf{Mid-to-late training:} Training task success increases steadily while test performance plateaus and slightly declines
    \item \textbf{Final performance:} 21.0\% success on training tasks vs. 13.2\% on test tasks, indicating a generalization gap
\end{itemize}

This behavior reflects a fundamental tension in meta-RL: optimizing for fast adaptation on the training distribution may reduce adaptability to unseen tasks.

\subsection{Task-Level Performance Analysis}

To understand which manipulation skills MAML learns effectively, we analyze final success rates per task.

\begin{figure*}[t]
    \centering
    \includegraphics[width=0.85\textwidth]{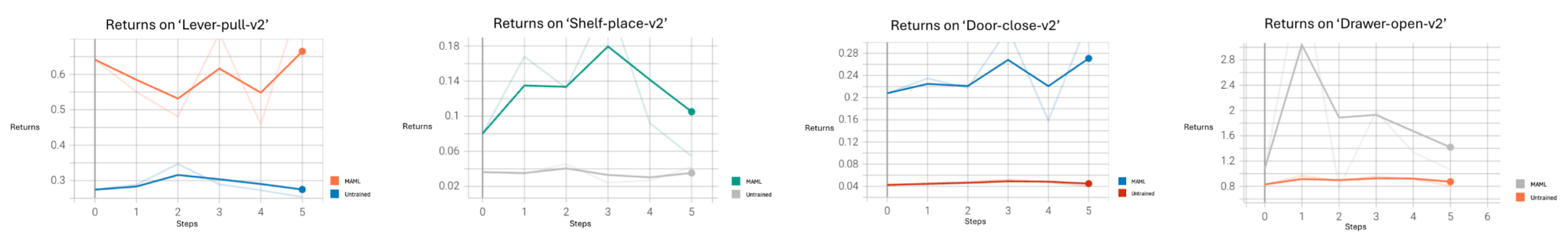}
    \caption{Final success rate (\%) by training task for MAML-TRPO on MetaWorld ML10. High variance across tasks, with some mastered (door-open) and others showing minimal learning (window-open).}
    \label{fig:training_tasks}
\end{figure*}

\textbf{Training Tasks (Figure \ref{fig:training_tasks}):}
\begin{itemize}
    \item \textbf{High success (>50\%):} door-open (80\%), drawer-close (55\%)
    \item \textbf{Moderate success (20-40\%):} button-press-topdown (30\%), sweep (25\%)
    \item \textbf{Low success (<10\%):} peg-insert-side (5\%), basketball (5\%), reach (3.33\%), window-open (0\%)
\end{itemize}

\begin{figure}[h]
    \centering
    \includegraphics[width=0.5\textwidth]{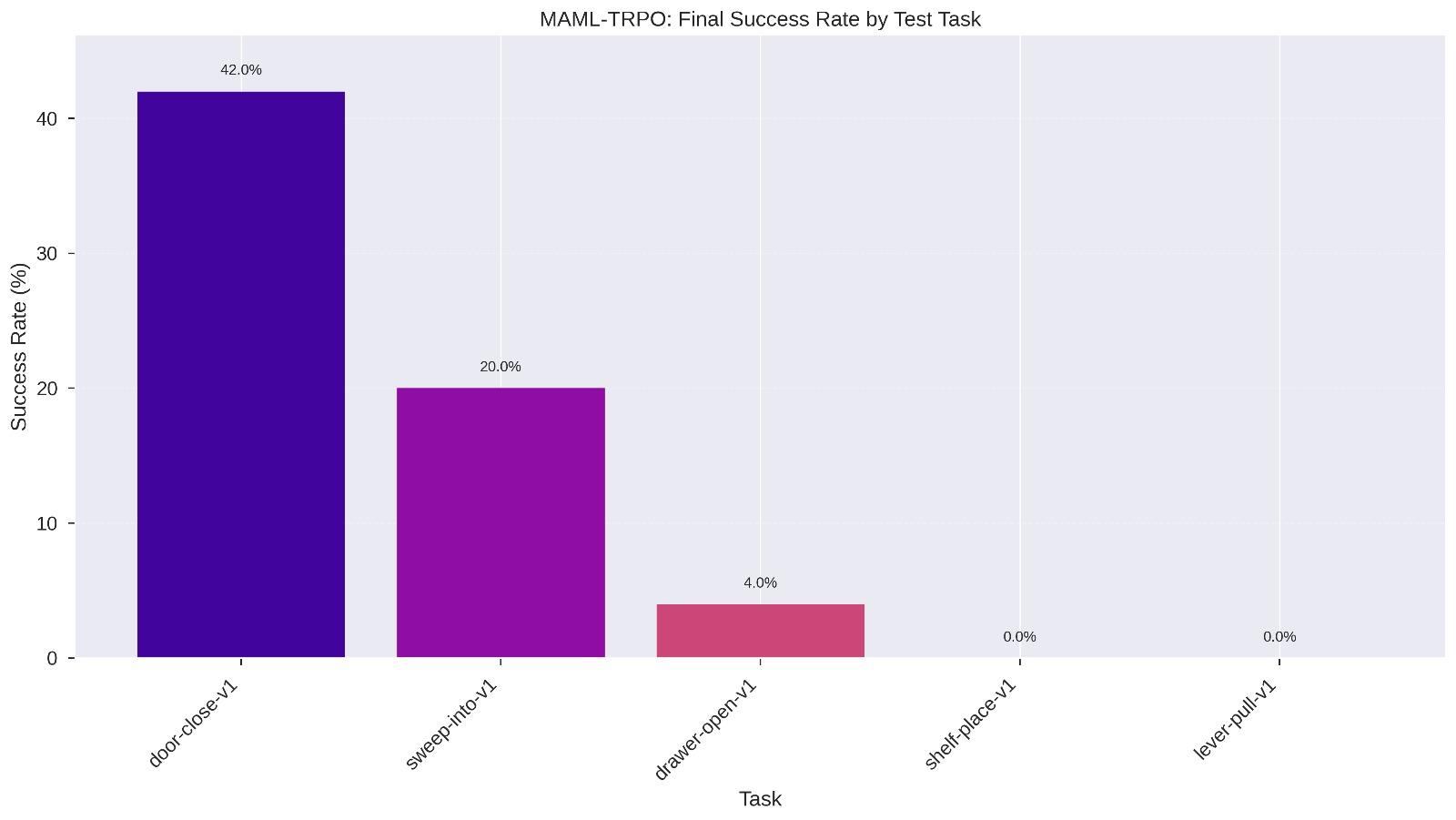}
    \caption{Final success rate (\%) by test task for MAML-TRPO. Results reflect MAML's ability to generalize to unseen tasks based on training experience.}
    \label{fig:test_tasks}
\end{figure}

\textbf{Test Tasks (Figure \ref{fig:test_tasks}):}
\begin{itemize}
    \item \textbf{Strong transfer:} door-close (42\%) benefits from similarity to door-open
    \item \textbf{Moderate transfer:} sweep-into (20\%) transfers from sweep
    \item \textbf{Poor transfer:} drawer-open (4\%), shelf-place (0\%), lever-pull (0\%)
\end{itemize}

Two key factors emerge:
\begin{enumerate}
    \item \textbf{Task similarity:} Transfer is strongest when training and test tasks share mechanical structure (e.g., door-open $\to$ door-close)
    \item \textbf{Task complexity:} Simpler tasks with sparse rewards (reaching, pressing) generalize better than complex sequential tasks (lever-pull, peg-insertion)
\end{enumerate}

\section{Discussion}

\subsection{Effectiveness of MAML for Diverse Manipulation}

Our results demonstrate that MAML-TRPO successfully learns a meta-initialization that enables rapid adaptation across diverse robotic manipulation tasks. The consistent gap between pre- and post-adaptation performance (Figure \ref{fig:training_loss}) confirms that a single gradient step meaningfully improves policy performance on new tasks.

However, the high variance in task-level success rates (Figures \ref{fig:training_tasks}, \ref{fig:test_tasks}) reveals limitations. MAML excels on structurally similar tasks but struggles with complex, sequential behaviors like lever-pulling or precise insertion. This suggests that purely gradient-based adaptation may be insufficient for tasks requiring substantial behavioral changes.

\subsection{The Generalization Gap}

The divergence between training and test task performance (Figure \ref{fig:generalization}) highlights a critical challenge in meta-RL: balancing adaptation speed on the training distribution with robustness to new tasks. This generalization gap likely arises from:

\begin{itemize}
    \item \textbf{Overfitting to task distribution:} The meta-optimizer specializes toward training tasks at the expense of broader adaptability
    \item \textbf{Limited task diversity:} While ML10 is more diverse than prior benchmarks, eight training tasks may be insufficient to prevent overfitting
    \item \textbf{Gradient-based adaptation limitations:} Tasks requiring qualitatively different behaviors may need more than gradient updates on policy parameters
\end{itemize}

Interestingly, test performance initially exceeds training performance, suggesting that early meta-initializations are more general. This could motivate early stopping or regularization strategies to preserve generalization.

\subsection{Multi-Step Adaptation}

The declining performance after the first gradient step on some tasks (Figure \ref{fig:gradient_steps}) reflects MAML's optimization for one-shot adaptation. While this is by design, it limits applicability when more fine-tuning is beneficial. Potential solutions include:
\begin{itemize}
    \item Meta-learning with multi-step inner loops
    \item Adaptive learning rates per task or step
    \item Hybrid approaches combining gradient-based and inference-based adaptation
\end{itemize}

\subsection{Comparison to Alternative Approaches}

Context-based meta-RL methods like PEARL \cite{rakelly2019efficient} use probabilistic task inference rather than gradient updates. These methods may handle diverse tasks more flexibly by explicitly modeling task identity. A direct comparison on ML10 would clarify the tradeoffs between gradient-based and inference-based adaptation.

Multi-task RL with modular architectures \cite{pong2019multi} offers another alternative, learning task-specific module compositions. Such approaches may scale better to task diversity but sacrifice the simplicity and generality of MAML.

\section{Conclusion}

This work evaluated Model-Agnostic Meta-Learning combined with Trust Region Policy Optimization on the MetaWorld ML10 benchmark for robotic manipulation. Our experiments demonstrate that MAML-TRPO learns an effective initialization for few-shot adaptation, with clear performance gains after a single gradient update. However, we identified several challenges:

\begin{itemize}
    \item High variance in task-level adaptation success, with performance ranging from 0\% to 80\%
    \item A generalization gap where test task performance plateaus while training performance improves
    \item Declining performance with additional gradient steps on some tasks
\end{itemize}

These findings suggest that while gradient-based meta-learning shows promise for robotic manipulation, current approaches struggle with task diversity and long-horizon behaviors. Future work should explore:

\begin{itemize}
    \item \textbf{Task-aware adaptation:} Incorporating task embeddings or context variables to guide adaptation before taking gradient steps
    \item \textbf{Structured policies:} Modular or hierarchical architectures better suited for diverse manipulation primitives
    \item \textbf{Hybrid meta-learning:} Combining gradient-based updates with inference-based task identification
    \item \textbf{Real-world evaluation:} Testing meta-learned policies on physical robots to assess sim-to-real transfer and robustness
\end{itemize}

Our implementation and experimental framework provide a foundation for these investigations. The MetaWorld ML10 benchmark, with its task diversity and standardized evaluation, represents a valuable testbed for advancing meta-learning in robotics.

\bibliographystyle{plain}

\begin{thebibliography}{9}

\bibitem{finn2017model}
Chelsea Finn, Pieter Abbeel, and Sergey Levine.
\newblock Model-Agnostic Meta-Learning for Fast Adaptation of Deep Networks.
\newblock In \textit{Proceedings of the 34th International Conference on Machine Learning (ICML 2017)}, pages 1126--1135, 2017.

\bibitem{ravi2017optimization}
Sachin Ravi and Hugo Larochelle.
\newblock Optimization as a Model for Few-Shot Learning.
\newblock In \textit{Proceedings of the 5th International Conference on Learning Representations (ICLR 2017)}, 2017.

\bibitem{antoniou2019train}
Antreas Antoniou, Harrison Edwards, and Amos Storkey.
\newblock How to Train Your MAML.
\newblock In \textit{Proceedings of the 7th International Conference on Learning Representations (ICLR 2019)}, 2019.

\bibitem{hospedales2020meta}
Timothy Hospedales, Antreas Antoniou, Paul Micaelli, and Amos Storkey.
\newblock Meta-Learning in Neural Networks: A Survey.
\newblock \textit{arXiv preprint arXiv:2004.05439}, 2020.

\bibitem{yu2019meta}
Tianhe Yu, Deirdre Quillen, Zhanpeng He, Ryan Julian, Avnish Narayan, Hayden Shively, Adithya Bellathur, Karol Hausman, Chelsea Finn, and Sergey Levine.
\newblock Meta-World: A Benchmark and Evaluation for Multi-Task and Meta Reinforcement Learning.
\newblock In \textit{Proceedings of the 3rd Conference on Robot Learning (CoRL 2019)}, 2019.

\bibitem{rakelly2019efficient}
Kate Rakelly, Aurick Zhou, Deirdre Quillen, Chelsea Finn, and Sergey Levine.
\newblock Efficient Off-Policy Meta-Reinforcement Learning via Probabilistic Context Variables.
\newblock In \textit{Proceedings of the 36th International International Conference on Machine Learning (ICML 2019)}, pages 5331--5340, 2019.

\bibitem{pong2019multi}
Vitchyr H. Pong, Murtaza Dalal, Steven Lin, Ashvin Nair, Shikhar Bahl, and Sergey Levine.
\newblock Multi-Task Reinforcement Learning with Soft Modularization.
\newblock In \textit{Proceedings of the 36th International Conference on Machine Learning (ICML 2019)}, pages 1071--1080, 2019.

\bibitem{schulman2015trust}
John Schulman, Sergey Levine, Pieter Abbeel, Michael Jordan, and Philipp Moritz.
\newblock Trust Region Policy Optimization.
\newblock In \textit{Proceedings of the 32nd International Conference on Machine Learning (ICML 2015)}, pages 1889--1897, 2015.

\end{thebibliography}

\end{document}